# The Entire Quantile Path of a Risk-Agnostic SVM Classifier

**Jin Yu**
Canberra Research Laboratory, NICTA
College of Engineering & Computer Science
Australian National University
Canberra, Australia
jin.yu@anu.edu.au

**S.V.N. Vishwanathan**        **Jian Zhang**
Department of Statistics
Purdue University
250 N University Street
West Lafayette, IN 47907-2066, USA
{vishy, jianzhan}@stat.purdue.edu

## Abstract

A quantile binary classifier uses the rule: Classify $\boldsymbol{x}$ as $+1$ if $P(Y = 1 | \boldsymbol{X} = \boldsymbol{x}) \geq \tau$, and as $-1$ otherwise, for a fixed quantile parameter $\tau \in [0, 1]$. It has been shown that Support Vector Machines (SVMs) in the limit are quantile classifiers with $\tau = \frac{1}{2}$. In this paper, we show that by using asymmetric cost of misclassification SVMs can be appropriately extended to recover, in the limit, the quantile binary classifier for any $\tau$. We then present a principled algorithm to solve the extended SVM classifier for *all* values of $\tau$ simultaneously. This has two implications: First, one can recover the entire conditional distribution $P(Y = 1 | \boldsymbol{X} = \boldsymbol{x}) = \tau$ for $\tau \in [0, 1]$. Second, we can build a risk-agnostic SVM classifier where the cost of misclassification need not be known apriori. Preliminary numerical experiments show the effectiveness of the proposed algorithm.

## 1 Introduction

Support Vector Machines (SVMs) have emerged as a popular tool for binary classification. Given a set of $n$ training instances $\boldsymbol{x}_i$ and their corresponding labels $y_i \in \{\pm 1\}$ the task of training a linear SVM classifier[1] can be cast as a regularized risk minimization problem:

$$\min J(\boldsymbol{w}) := \frac{\lambda}{2} \|\boldsymbol{w}\|^2 + \frac{1}{n} \sum_{i=1}^{n} l(\boldsymbol{w}^\top \boldsymbol{x}_i, y_i), \quad (1)$$

where $l$ is the so-called hinge loss defined by

$$l(\boldsymbol{w}^\top \boldsymbol{x}, y) := \max(0, 1 - y \, \boldsymbol{w}^\top \boldsymbol{x}). \quad (2)$$

---

[1]For ease of exposition we will stick to linear SVMs, although all our results extend to the non-linear case where one maps the data into an RKHS $\mathcal{H}$ via the map $\boldsymbol{x} \to \phi(\boldsymbol{x})$ and uses the kernel $k(\boldsymbol{x}, \boldsymbol{x}') = \langle \phi(\boldsymbol{x}), \phi(\boldsymbol{x}') \rangle_{\mathcal{H}}$.

It is obvious that the hinge loss $l(\cdot)$ is non-negative and convex in $\boldsymbol{w}$. The loss function measures the discrepancy between $y$ and the prediction given by $\text{sign}(\boldsymbol{w}^\top \boldsymbol{x})$, while the $L_2$ regularizer with regularization constant $\lambda > 0$ controls the complexity of the solution $\boldsymbol{w}$. It has been shown (Lin, 2002) that the minimizer of the hinge loss is exactly $\text{sign}(\eta(\boldsymbol{x}) - 1/2)$, where $\eta(\boldsymbol{x}) = P(Y = 1 | \boldsymbol{X} = \boldsymbol{x})$ is the probability of $Y = 1$ conditioned on $\boldsymbol{X} = \boldsymbol{x}$. Thus the SVM solution should approach the Bayes rule as the sample size gets large with appropriately chosen function class.

One problem of the standard SVM is that even though we can use the resulting SVM classifier $\hat{\boldsymbol{w}} = \arg\min_{\boldsymbol{w}} J(\boldsymbol{w})$ to classify a new observation $\boldsymbol{x}$, the prediction $\boldsymbol{w}^\top \boldsymbol{x}$ does not have probabilistic interpretation. More importantly, the SVM classification results cannot be directly applied to situations where the misclassification cost is asymmetric, i.e. when the cost of a false positive error is different from that of a false negative error. To address such a problem, several methods have been proposed to convert the SVM output into well-calibrated probabilistic scores, such as (Platt, 2000). However, such methods either rely on parametric assumption or lack theoretical justification about the transformed scores.

We instead aim to estimate the quantity $\text{sign}(\eta(\boldsymbol{x}) - \tau)$, which we call the quantile classification rule with $\tau \in [0, 1]$ as the quantile parameter. It can be shown that $\text{sign}(\eta(\boldsymbol{x}) - \tau)$ is the minimizer of the asymmetric hinge loss, which assigns different costs to false positive and false negative errors. The SVM formulation with the asymmetric hinge loss can be defined as

$$\min \frac{\lambda}{2} \|\boldsymbol{w}\|^2 + \sum_{i=1}^{n} c_{y_i}^{\tau} \max(0, 1 - y_i \, \boldsymbol{w}^\top \boldsymbol{x}_i). \quad (3)$$

Here, $c_{y_i}^{\tau}$ controls the two types of misclassification cost, and for reasons that will become apparent

shortly, we set

$$c_{y_i}^{\tau} := \begin{cases} [2(1-\tau)]/n & \text{if } y_i = +1, \\ (2\tau)/n & \text{if } y_i = -1. \end{cases} \qquad (4)$$

for $\tau \in [0, 1]$. Note that when $\tau = \frac{1}{2}$, $i.e.$, the misclassification costs are symmetric, we recover (1).

There are many natural applications where the cost of misclassification is not known in advance until the classifier is deployed. As an illustrative example consider the problem of spam detection (also see Figure 1). Given training emails the learning task is to distinguish between spam and non-spam emails. The tolerance of a user to spam is influenced by various factors. For instance, a busy professor might have a very low tolerance for spam. In other words, he/she might not mind losing a few genuine emails as long as all spam emails are kept out of his/her inbox. On the other hand, a not-so-busy graduate student might not mind a few spam emails as long as genuine emails are not lost in the junk mail folder. In such cases, the brute force approach of training a classifier for every user preference is both tedious and time consuming. Furthermore, one needs to train a new classifier for every user preference.

In this paper we present a principled algorithm to solve (3) for *all* values of $\tau$ simultaneously by utilizing the fact that the solution path is piecewise linear as a function of the quantile parameter $\tau \in [0, 1]$. In other words, once our classifier is trained, we can recover the solution for any $\tau$ efficiently. Consequently, our classifier is risk agnostic. Furthermore, we show that (3) is an instance of the quantile classification problem, with $\tau$ being the quantile parameter.

The rest of the paper is organized as follows: In Section 2 we establish the connection between the asymmetric cost SVM and the quantile classification rule. In Section 3 we formulate the dual of (3). In Section 4 we describe the proposed algorithm and its worst case time complexity analysis. We discuss some related work in Section 5. Numerical experiments are presented in Section 6, and we conclude the paper with an outlook and discussion in Section 7.

## 2 Statistical Underpinnings

Let $(\boldsymbol{X}, Y)$ be a pair of random variables with training instances $\boldsymbol{X} \in \mathcal{X}$ and labels $Y \in \{\pm 1\}$. For any realization $\boldsymbol{x}$ of $\boldsymbol{X}$, denote the conditional probability $P(Y = 1 | \boldsymbol{X} = \boldsymbol{x})$ as $\eta(\boldsymbol{x})$. Furthermore, let $C_+$ (resp. $C_-$) denote the cost of misclassifying a $\boldsymbol{x}$ labeled as $+1$ (resp. $-1$). The cost sensitive classification risk of

a decision function $g : \mathcal{X} \to \{\pm 1\}$ is defined as

$$\begin{aligned} R(g) := &\ C_+ P(Y = -1, g(\boldsymbol{X}) = 1) \qquad (5) \\ &+ C_- P(Y = 1, g(\boldsymbol{X}) = -1). \end{aligned}$$

The following lemma follows from elementary Bayes decision theory (see $e.g.$, Section 2.2 of Duda et al., 2001).

**Lemma 2.1** *For any decision function* $g$

$$R(\text{sign}(\eta(\boldsymbol{x}) - \tau)) \leq R(g),$$

*where* $\tau = \frac{C_+}{C_+ + C_-} \in [0, 1]$.

Lemma 2.1 says that when the misclassification cost is asymmetric, the classifier which leads to the minimum risk should take the form $\text{sign}(\eta(\boldsymbol{x}) - \tau)$ where $\tau$ only depends on the ratio of the misclassification costs $C_+/C_-$. The following lemma, whose proof can be found in Appendix A, shows that $\text{sign}(\eta(\boldsymbol{x}) - \tau)$ is the minimizer of the asymmetric hinge loss.

**Lemma 2.2** *For any* $\tau \in [0, 1]$, *the minimizer* $f^*$ *of*

$$\mathbb{E}_{\boldsymbol{X}, Y} \left[ ((1 + Y)/2 - \tau Y) (1 - Y f(\boldsymbol{X}))_+ \right] \qquad (6)$$

*takes the form* $f^*(\boldsymbol{x}) = \text{sign}(\eta(\boldsymbol{x}) - \tau)$.

In the infinite sample case, if we let $\lambda \to 0$ as $n \to \infty$ in (3), it is easy to see that the regularized risk converges to (6). Therefore, Lemma 2.2 implies that the estimator obtained by minimizing the regularized risk (3) is risk consistent.

The above observation has a number of consequences. First, it shows that the usual SVM with the symmetric hinge loss (2) estimates $P(Y = 1 | \boldsymbol{X} = \boldsymbol{x}) = \frac{1}{2}$, a well known result (see $e.g.$, Lin, 2002; Sollich, 2002). It also shows that the SVM with the asymmetric hinge loss (4) is essentially a quantile estimator. This result has been hinted many times (e.g. Grandvalet et al., 2006), but to the best of our knowledge, not proven rigorously.

## 3 Dual Formulation

Similar to the case of standard SVM, we can rewrite (3) as a constrained optimization problem:

$$\min \quad \frac{\lambda}{2} \|\boldsymbol{w}\|^2 + \sum_{i=1}^{n} c_{y_i}^{\tau} \xi_i \qquad (7)$$

$$\text{s.t.} \quad \xi_i \geq 0, \quad \xi_i \geq 1 - y_i \boldsymbol{w}^\top \boldsymbol{x}_i, \quad \forall i$$

allows us to derive its dual as:

$$\min \quad D(\boldsymbol{\alpha}) := \frac{1}{2\lambda} \boldsymbol{\alpha}^\top Q \boldsymbol{\alpha} - \boldsymbol{\alpha}^\top \mathbf{1} \qquad (8)$$

$$\text{s.t.} \quad 0 \leq \alpha_i \leq c_{y_i}^{\tau}, \quad \forall i.$$

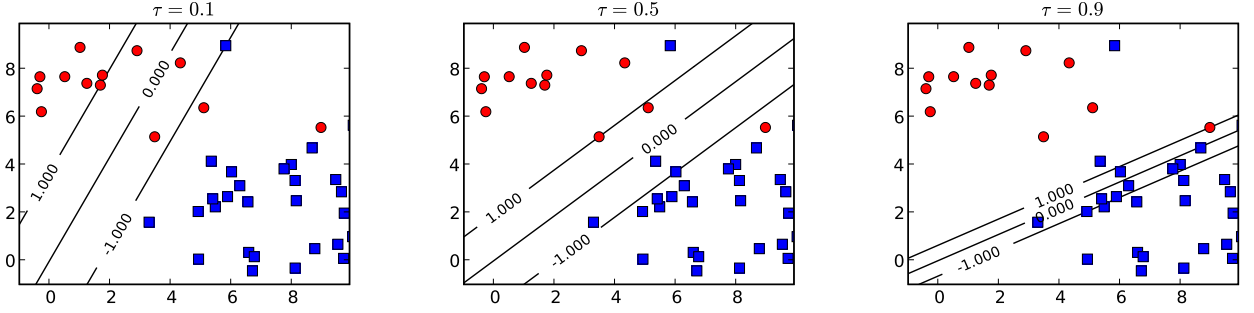

Figure 1: The effect of $\tau$ on the decision boundary of a binary quantile classifier on synthetically generated two dimensional data. When $\tau = \frac{1}{2}$ (middle) the cost of misclassification is symmetric and therefore the resulting classifier is well balanced. For low values of $\tau$ (left), the classifier classifies all blue squares correctly even at the cost of misclassifying a few red dots. A converse effect is observed for high values of $\tau$ (right).

where $\mathbf{1}$ is a vector of all ones and $Q_{ij} = y_i y_j \boldsymbol{x}_i^\top \boldsymbol{x}_j$. Let $\boldsymbol{\alpha}^\tau$ denote the optimal solution to (8) for a given $\tau$. Then the primal solution $\boldsymbol{w}^\tau$ can be recovered via the primal-dual connection:

$$\boldsymbol{w}^\tau = \frac{1}{\lambda} \sum_{i=1}^n \alpha_i^\tau y_i \boldsymbol{x}_i. \qquad (9)$$

The dual problem is a quadratic problem with box constrains, which can be solved by various optimization techniques (see *e.g.,* Byrd et al., 1995; Moré and Toraldo, 1989)

Define $\nabla_i D(\boldsymbol{\alpha}^\tau)$ as the $i^{\text{th}}$ element of the gradient:

$$\nabla D(\boldsymbol{\alpha}^\tau) = \frac{1}{\lambda} Q \boldsymbol{\alpha}^\tau - \mathbf{1}; \qquad (10)$$

and let $\mathcal{L}$, $\mathcal{M}$, and $\mathcal{R}$ index entries of the dual solution $\boldsymbol{\alpha}^\tau$ such that

$$\mathcal{L} := \{i : \nabla_i D(\boldsymbol{\alpha}^\tau) < 0\} = \{i : y_i \boldsymbol{w}^{\tau\top} \boldsymbol{x}_i < 1\},$$
$$\mathcal{M} := \{i : \nabla_i D(\boldsymbol{\alpha}^\tau) = 0\} = \{i : y_i \boldsymbol{w}^{\tau\top} \boldsymbol{x}_i = 1\}, \qquad (11)$$
$$\mathcal{R} := \{i : \nabla_i D(\boldsymbol{\alpha}^\tau) > 0\} = \{i : y_i \boldsymbol{w}^{\tau\top} \boldsymbol{x}_i > 1\},$$

where the connection with the primal parameter $\boldsymbol{w}^\tau$ is made via (9). It is easy to find that $\mathcal{L}$, $\mathcal{M}$, and $\mathcal{R}$ in fact index the data $\boldsymbol{x}_i$ which are in error, on the margin, and well-classified, respectively. Furthermore, it follows from the KKT conditions (see Appendix B) that the optimal dual solution $\boldsymbol{\alpha}^\tau$ satisfies

$$\alpha_i^\tau = \begin{cases} c_{y_i}^\tau & \text{if } i \in \mathcal{L}, \\ [0, c_{y_i}^\tau] & \text{if } i \in \mathcal{M}, \\ 0 & \text{if } i \in \mathcal{R}. \end{cases} \qquad (12)$$

Given index sets, $\mathcal{I}$ and $\mathcal{J}$, let $\boldsymbol{\alpha}_{\mathcal{I}}^\tau$ be a vector of $\alpha_i^\tau$ with $i \in \mathcal{I}$ and $Q_{\mathcal{I}\mathcal{J}}$ a submatrix of $Q$ taking entries $Q_{ij}$ with $i \in \mathcal{I}$ and $j \in \mathcal{J}$. Then, we can define $\mathcal{L}_{+1} := \mathcal{L} \cap \{i : y_i = +1\}$ and $\mathcal{L}_{-1} := \mathcal{L} \setminus \mathcal{L}_{+1}$, and use (12) to decompose $\boldsymbol{\alpha}^\tau$ into $[\boldsymbol{\alpha}_{\mathcal{L}_+}^{\tau\top}, \boldsymbol{\alpha}_{\mathcal{L}_-}^{\tau\top}, \boldsymbol{\alpha}_{\mathcal{M}}^{\tau\top}, \mathbf{0}^\top]^\top$, where $\mathbf{0}$ is a vector of all zeros. Using (10), we can get a decomposed view of $\nabla D(\boldsymbol{\alpha}^\tau)$:

$$[\nabla_{\mathcal{L}} D(\boldsymbol{\alpha}^\tau)^\top, \nabla_{\mathcal{M}} D(\boldsymbol{\alpha}^\tau)^\top, \nabla_{\mathcal{R}} D(\boldsymbol{\alpha}^\tau)^\top]^\top := \quad (13)$$
$$\frac{1}{\lambda} \begin{bmatrix} c_{+1}^\tau Q_{\mathcal{L}\mathcal{L}_{+1}} \mathbf{1} + c_{-1}^\tau Q_{\mathcal{L}\mathcal{L}_{-1}} \mathbf{1} + Q_{\mathcal{L}\mathcal{M}} \boldsymbol{\alpha}_{\mathcal{M}}^\tau \\ c_{+1}^\tau Q_{\mathcal{M}\mathcal{L}_{+1}} \mathbf{1} + c_{-1}^\tau Q_{\mathcal{M}\mathcal{L}_{-1}} \mathbf{1} + Q_{\mathcal{M}\mathcal{M}} \boldsymbol{\alpha}_{\mathcal{M}}^\tau \\ c_{+1}^\tau Q_{\mathcal{R}\mathcal{L}_{+1}} \mathbf{1} + c_{-1}^\tau Q_{\mathcal{R}\mathcal{L}_{-1}} \mathbf{1} + Q_{\mathcal{R}\mathcal{M}} \boldsymbol{\alpha}_{\mathcal{M}}^\tau \end{bmatrix} - \mathbf{1}$$

Since $\nabla_{\mathcal{M}} D(\boldsymbol{\alpha}^\tau) = \mathbf{0}$ by the definition (11) of $\mathcal{M}$, we can use (13) to get a closed form representation of $\boldsymbol{\alpha}_{\mathcal{M}}^\tau$:

$$\boldsymbol{\alpha}_{\mathcal{M}}^\tau = Q_{\mathcal{M}\mathcal{M}}^{-1} [\lambda \mathbf{1} - c_{+1}^\tau Q_{\mathcal{M}\mathcal{L}_{+1}} \mathbf{1} - c_{-1}^\tau Q_{\mathcal{M}\mathcal{L}_{-1}} \mathbf{1}]. \quad (14)$$

Note that $Q_{\mathcal{M}\mathcal{M}}^{-1}$ does not exist when $Q_{\mathcal{M}\mathcal{M}}$ is rank deficient. Nevertheless, standard optimization techniques, such as the conjugate gradient method (Nocedal and Wright, 1999), should always recover $\boldsymbol{\alpha}_{\mathcal{M}}^\tau$ as a solution to a linear system.

An important observation from (4) is that the upper bound $c_{y_i}^\tau$ only changes linearly with $\tau$: As we increase the quantile parameter $\tau$ to $\tau + \epsilon$, we have

$$c_{y_i}^{\tau+\epsilon} = c_{y_i}^\tau - y_i \, \Delta c^\epsilon, \quad \text{where } \Delta c^\epsilon := \frac{2\epsilon}{n}. \quad (15)$$

Assume an $\epsilon$ deviation from $\tau$ does not change the index sets defined in (11), then (14) still holds for $\boldsymbol{\alpha}_{\mathcal{M}}^{\tau+\epsilon}$. Therefore, we can use (15) to expand it as:

$$\boldsymbol{\alpha}_{\mathcal{M}}^{\tau+\epsilon} = \boldsymbol{\alpha}_{\mathcal{M}}^\tau + \Delta c^\epsilon \, \Delta \boldsymbol{\alpha}_{\mathcal{M}}, \quad \text{where} \quad (16)$$
$$\Delta \boldsymbol{\alpha}_{\mathcal{M}}^\tau := Q_{\mathcal{M}\mathcal{M}}^{-1} [Q_{\mathcal{M}\mathcal{L}_{+1}} \mathbf{1} - Q_{\mathcal{M}\mathcal{L}_{-1}} \mathbf{1}].$$

The optimality condition (12) then allows us to recover $\boldsymbol{\alpha}^{\tau+\epsilon}$ from $\boldsymbol{\alpha}^\tau$ via:

$$\begin{bmatrix} \boldsymbol{\alpha}_{\mathcal{L}_{+1}}^{\tau+\epsilon} \\ \boldsymbol{\alpha}_{\mathcal{L}_{-1}}^{\tau+\epsilon} \\ \boldsymbol{\alpha}_{\mathcal{M}}^{\tau+\epsilon} \\ \boldsymbol{\alpha}_{\mathcal{R}}^{\tau+\epsilon} \end{bmatrix} = \begin{bmatrix} \boldsymbol{\alpha}_{\mathcal{L}_{+1}}^\tau - \Delta c^\epsilon \\ \boldsymbol{\alpha}_{\mathcal{L}_{-1}}^\tau + \Delta c^\epsilon \\ \boldsymbol{\alpha}_{\mathcal{M}}^\tau + \Delta c^\epsilon \, \Delta \boldsymbol{\alpha}_{\mathcal{M}} \\ \mathbf{0} \end{bmatrix}. \quad (17)$$

**Proposition 3.1** *For the dual of the quantile classification problem* (8), *there exists a set of quantiles* $\{\tau_k\}_{k=1}^K, \tau_k \in [0,1]$, *such that we can find a solution path* $\boldsymbol{\alpha}^\tau$ *that is continuous in* $\tau$, *and linear in* $\tau, \forall \tau \in (\tau_k, \tau_{k+1})$.

See Appendix C for the proof of Proposition 3.1.

Proposition 3.1 shows that $\boldsymbol{\alpha}^\tau$ is piecewise linear in $\tau$. Using (13) and (17), we can see that the gradient $\nabla D(\boldsymbol{\alpha}^\tau)$ has the same property. In particular, $\forall \epsilon \in (0, \tau_{k+1} - \tau_k)$, we have $\nabla_{\mathcal{M}} D(\boldsymbol{\alpha}^{(\tau_k+\epsilon)}) = \mathbf{0}$ and

$$\nabla_{\mathcal{N}} D(\boldsymbol{\alpha}^{(\tau_k+\epsilon)}) = \nabla_{\mathcal{N}} D(\boldsymbol{\alpha}^{\tau_k})$$
$$+ \frac{\Delta c^\epsilon}{\lambda} \left[ Q_{\mathcal{N}\mathcal{L}-1} \mathbf{1} - Q_{\mathcal{N}\mathcal{L}+1} \mathbf{1} + Q_{\mathcal{N}\mathcal{M}} \; \Delta\boldsymbol{\alpha}_{\mathcal{M}}^{\tau_k} \right], \quad (18)$$

where $\mathcal{M}$ is the margin index set associated with $\boldsymbol{\alpha}^{\tau_k+\epsilon}$ and $\mathcal{N} := \mathcal{L} \cup \mathcal{R}$ is the complement set of $\mathcal{M}$.

## 4 Finding the Dual Solution Path

It follows from Proposition 3.1 that if we can find a set of quantile parameters: $\mathcal{K} := \{\tau_k\}_{k=1}^K$, that divide the interval $[0,1]$ into regions so that within these regions $\boldsymbol{\alpha}^\tau$ changes linearly with $\tau$, *i.e.*, the index sets: $\mathcal{L}, \mathcal{M}$, and $\mathcal{R}$ remain fixed. Then we can quickly recover $\boldsymbol{\alpha}^\tau$ for any value of $\tau$ from a $\boldsymbol{\alpha}^{\tau_k}$ via (17). In what follows we present our algorithm (Algorithm 1) that is able to identify all $\tau_k$, which we call *kinks*.

### 4.1 The Algorithm

Our goal is to construct a sorted list of kinks $\{\tau_k\}_{k=1}^K$, at which one of the following events happens:

1. Elements in $\mathcal{N}$, *i.e.*, *not* in $\mathcal{M}$, move to $\mathcal{M}$,
2. Elements in $\mathcal{M}$ move to $\mathcal{L}$,
3. Elements in $\mathcal{M}$ move to $\mathcal{R}$.

To this end, our algorithm starts with $\tau = 0$, and then moves forward toward $\tau = 1$ to identify all values of $\tau$ that alter the membership of an index.

Given a quantile parameter $\tau_k$, its corresponding optimal dual solution $\boldsymbol{\alpha}^{\tau_k}$, and the associated index sets $\mathcal{L}, \mathcal{M}$, and $\mathcal{R}$, we know from the definition (11) that $\nabla_i D(\boldsymbol{\alpha}^{\tau_k}) \neq 0, \forall i \in \mathcal{N}$. This means that Event 1 happens when an $\epsilon > 0$ deviation from $\tau_k$ just turns a nonzero element of the gradient to zero, *i.e.*, $\nabla_i D(\boldsymbol{\alpha}^{\tau_k+\epsilon}) = 0, i \in \mathcal{N}$. We immediately see from (18) that the deviation that leads to Event 1 is:

$$\epsilon^{\text{to}\,\mathcal{M}} = \min\{\epsilon_i : \epsilon_i > 0\}_{i \in \mathcal{N}}, \quad \text{where} \quad (19)$$
$$\epsilon_i := \frac{n}{2} \left[ \frac{-\lambda \nabla_i D(\boldsymbol{\alpha}^{\tau_k})}{Q_{i\,\mathcal{L}-1} \mathbf{1} - Q_{i\,\mathcal{L}+1} \mathbf{1} + Q_{i\,\mathcal{M}} \; \Delta\boldsymbol{\alpha}_{\mathcal{M}}^{\tau_k}} \right].$$

We know from the optimality condition (12) that an index $i$ from $\mathcal{M}$ is just about to move into $\mathcal{L}$ (Event 2), when $\alpha_i^{(\tau_k+\epsilon)} = c_{y_i}^{(\tau_k+\epsilon)}$. Expanding both sides of the last equation, using (15) and (16), shows that $\epsilon$ satisfies

$$\alpha_i^{\tau_k} + \frac{2\epsilon}{n} \; \Delta\alpha_i^{\tau_k} = c_{y_i}^{\tau_k} - y_i \frac{2\epsilon}{n}, \quad i \in \mathcal{M}. \quad (20)$$

Here care must be taken when $\alpha_i^{\tau_k} = c_{y_i}^{\tau_k}$, $i \in \mathcal{M}$, *i.e.*, $\alpha_i^{\tau_k}$ is on the boundary between $\mathcal{L}$ and $\mathcal{M}$. In this case (20) can be reduced to $\epsilon\Delta\alpha_i^{\tau_k} = -\epsilon y_i$; and if $\Delta\alpha_i^{\tau_k} > -y_i$, then an arbitrarily small $\epsilon > 0$ will cause $\alpha_i^{(\tau_k+\epsilon)} > c_{y_i}^{(\tau_k+\epsilon)}$, *i.e.*, pushing the index $i$ out toward $\mathcal{L}$. Taking this boundary case into consideration, we determine a candidate $\epsilon$ using the following criteria:

$$\epsilon^{\text{to}\,\mathcal{L}} = \min\{\epsilon_i : \epsilon_i \geq 0\}_{i \in \mathcal{M}}, \quad \text{where} \quad (21)$$
$$\epsilon_i := \begin{cases} 0 & \text{if } \alpha_i^{\tau_k} = c_{y_i}^{\tau_k} \; \& \; \Delta\alpha_i^{\tau_k} > -y_i, \\ +\infty & \text{if } \alpha_i^{\tau_k} = c_{y_i}^{\tau_k} \; \& \; \Delta\alpha_i^{\tau_k} \leq -y_i, \\ \frac{n}{2}(c_{y_i}^{\tau_k} - \alpha_i^{\tau_k})/(\Delta\alpha_i^{\tau_k} + y_i) & \text{otherwise.} \end{cases}$$

If $\epsilon^{\text{to}\,\mathcal{L}} = 0$, we treat $\tau_k$ as a kink, and update the index sets accordingly:

$$\mathcal{M} \leftarrow \mathcal{M} \backslash \{i^{\text{to}\,\mathcal{L}}\}, \quad \mathcal{L} \leftarrow \mathcal{L} \cup \{i^{\text{to}\,\mathcal{L}}\}, \quad (22)$$
$$\text{where} \quad i^{\text{to}\,\mathcal{L}} = \{i : i \in \mathcal{M}, \epsilon_i = \epsilon^{\text{to}\,\mathcal{L}}\},$$

such that the updated index sets coincide with the index sets of the optimal dual solution $\boldsymbol{\alpha}^\tau, \forall \tau \in (\tau_k, \tau_{k+1}), \tau_{k+1}$ being the next kink.

Similarly, to detect Event 3, we find $\epsilon$ that satisfies $\alpha_i^{\tau_k+\epsilon} = 0, \; \forall i \in \mathcal{M}$, and isolate $\alpha_i^{\tau_k} = 0$ case for special treatment.

$$\epsilon^{\text{to}\,\mathcal{R}} = \min\{\epsilon_i : \epsilon_i \geq 0\}_{i \in \mathcal{M}}, \quad \text{where} \quad (23)$$
$$\epsilon_i := \begin{cases} 0 & \text{if } \alpha_i^{\tau_k} = 0 \; \& \; \Delta\alpha_i^{\tau_k} < 0, \\ +\infty & \text{if } \alpha_i^{\tau_k} = 0 \; \& \; \Delta\alpha_i^{\tau_k} \geq 0, \\ \frac{n}{2}(-\alpha_i^{\tau_k})/(\Delta\alpha_i^{\tau_k}) & \text{otherwise.} \end{cases}$$

In the case where $\epsilon^{\text{to}\,\mathcal{R}} = 0$, we should recognize $\tau_k$ as a kink, and shift the corresponding index from $\mathcal{M}$ to $\mathcal{R}$. See Algorithm 1 for detailed implementation.

### 4.2 Complexity Analysis

The time complexity of Algorithm 1 is dominated by the calculation of $\Delta\boldsymbol{\alpha}_{\mathcal{M}}^\tau$ (16), which involves solving a linear system of size $|\mathcal{M}|$. A standard solver such as the conjugate gradient method converges to the solution of such a linear system in at most $O(r|\mathcal{M}|^2)$ time, $r$ being the rank of $Q_{\mathcal{M}\mathcal{M}}$. Having computed $\Delta\boldsymbol{\alpha}_{\mathcal{M}}^\tau$, the main cost of finding $\epsilon^{\text{to}\,\mathcal{M}}$ (19) is the $O(n|\mathcal{N}|)$ cost of matrix-vector multiplication; and the

**Algorithm 1** DUAL PATH FINDING (DPF)

1: **input** data $\{(\boldsymbol{x}_i, y_i)\}_{i=1}^n$ and regularizer $\lambda$
2: **output** sorted list of kinks $\tau_k$, corresponding dual solution $\boldsymbol{\alpha}^{\tau_k}$ and index sets.
3: $\boldsymbol{\alpha}^0 = \operatorname{argmin} D(\boldsymbol{\alpha})$, s.t. $0 \leq \alpha_i \leq c_{y_i}^0$, $\forall i$
4: construct $\mathcal{L}$, $\mathcal{M}$ and $\mathcal{R}$ for $\boldsymbol{\alpha}^0$ via (11)
5: calculate $\nabla D(\boldsymbol{\alpha}^0)$ and $\Delta\boldsymbol{\alpha}_\mathcal{M}^0$
6: $\tau \leftarrow 0$ and $\mathcal{K} \leftarrow \{(0, \boldsymbol{\alpha}_\mathcal{M}^\tau, \mathcal{L}, \mathcal{M})\}$
7: **while** $\tau < 1$ **do**
8: $\quad \epsilon \leftarrow \min\{\epsilon^{\text{to}\,\mathcal{M}}(19), \ \epsilon^{\text{to}\,\mathcal{L}}(21), \ \epsilon^{\text{to}\,\mathcal{R}}(23)\}$
9: $\quad$ update $\nabla D(\boldsymbol{\alpha}^\tau)$ to $\nabla D(\boldsymbol{\alpha}^{\tau+\epsilon})$ via (18)
10: $\quad$ update $\boldsymbol{\alpha}^\tau$ to $\boldsymbol{\alpha}^{\tau+\epsilon}$ via (17)
11: $\quad$ adjust index sets according to the event that $\epsilon$ triggers, *e.g.*, if $\epsilon = \epsilon^{\text{to}\,\mathcal{L}}$, apply (22)
12: $\quad \tau \leftarrow \tau + \epsilon$
13: $\quad$ calculate $\Delta\boldsymbol{\alpha}_\mathcal{M}^\tau$
14: $\quad$ **if** $\epsilon = 0$ **then**
15: $\quad\quad$ overwrite the last element of $\mathcal{K}$ with $(\tau, \boldsymbol{\alpha}_\mathcal{M}^\tau, \mathcal{L}, \mathcal{M})$        (*cf.* discussion in Sec. 4.1)
16: $\quad$ **else**
17: $\quad\quad \mathcal{K} \leftarrow \mathcal{K} \cup \{(\tau, \boldsymbol{\alpha}_\mathcal{M}^\tau, \mathcal{L}, \mathcal{M})\}$
18: $\quad$ **end if**
19: **end while**
20: **return** $\mathcal{K}$

time required to find $\epsilon^{\text{to}\,\mathcal{L}}$ (21) and $\epsilon^{\text{to}\,\mathcal{R}}$ (23) is linear in $|\mathcal{M}|$. Once we find all the kinks, we can recover $\boldsymbol{\alpha}^\tau$ for any $\tau \in [0,1]$ via (17) in $O(n)$ time by noting that $\boldsymbol{\alpha}_\mathcal{M}^\tau = \boldsymbol{\alpha}_\mathcal{M}^{\tau_k} + \Delta c^{(\tau-\tau_k)} \Delta\boldsymbol{\alpha}_\mathcal{M}^{\tau_k}$, where $\Delta\boldsymbol{\alpha}_\mathcal{M}^{\tau_k} = (\boldsymbol{\alpha}_\mathcal{M}^{\tau_{k+1}} - \boldsymbol{\alpha}_\mathcal{M}^{\tau_k})/\Delta c^{(\tau_{k+1}-\tau_k)}$ and $\mathcal{M}$ is associated with $\boldsymbol{\alpha}^{\tau_k}$.

In term of memory requirement, saving kink information at Step 17 of Algorithm 1 requires $O(|\mathcal{M}|)$ space for $\boldsymbol{\alpha}_\mathcal{M}^\tau$; and after the initial $O(n)$ cost of saving the entire $\mathcal{L}$ and $\mathcal{M}$ sets, we only need to keep track of the indices that move into or out of the two sets to recover them from their initial copy. We use the $Q$ matrix in our calculation, *e.g.*, (19). The $Q$ matrix is usually dense; and caching it requires $O(n^2)$ space. This can be prohibitively expensive. However, noticing that $Q = (YX)(X^\top Y)$, where $Y$ is a diagonal matrix with labels $y_i$ on its diagonal and $X = [\boldsymbol{x}_1^\top, \cdots, \boldsymbol{x}_n^\top] \in \mathbb{R}^{n \times d}$ is a feature matrix, we can instead cache $YX$, which is often very sparse. But, constructing $Q$ from $YX$ at each step can be computationally expensive. Fortunately, since only the product of $Q$ with a vector $\boldsymbol{v}$ is needed for our calculation, we can calculate it as $Q\boldsymbol{v} = YX(X^\top Y\boldsymbol{v})$ to leverage fast sparse matrix-vector product, and hence reduce the computational overhead. Although we do not have a formal bound on the size of $|\mathcal{K}|$, our experiments show that it never exceeds $O(n \log n)$.

# 5  Related Work

Perhaps the closest in spirit to our paper is the work of Hastie et al. (2004), who studied the influence of the regularization constant $\lambda$ on the generalization performance of a binary SVM. They showed that solutions to a SVM training problem is a piecewise linear function of $\lambda$. Based on this observation, they proposed an algorithm that finds SVM solutions for all values of $\lambda$. The regularization constant controls the balance between the regularization term and the empirical risk in the objective function (1) to prevent a classifier from overfitting the training data. Therefore, it plays an important role in improving prediction accuracy on unseen data. The effect of $\tau$ on the behaviour of a SVM classifier is fundamentally different from that of $\lambda$ in a sense that $\tau$ determines the trade-off between the true positive rate (TPR) and the true negative rate (TNR) of a classifier by assigning asymmetric costs to false positive and false negative predictions. In applications where an appropriate balance between TPR and TNR is considered to be more important than prediction accuracy, *e.g.*, in medical diagnosis, using a quantile classifier (3) with adjustable $\tau$ may be more desirable.

Although SVM classifiers with built-in asymmetric misclassification costs have been applied to classification problems that are characterized by highly skewed training data and to problems arisen from medical diagnosis (Veropoulos et al., 1999; Morik et al., 1999; Grandvalet et al., 2006), no rigorous statistical properties were established. The misclassification cost is commonly chosen to reflect label proportions of training data or the ratio of false positive cost to false negative cost. From the optimization viewpoint training a SVM with asymmetric costs is very similar to the standard SVM training problem. Hence, optimization software such as SVM$^{perf}$ (Joachims, 2006) and LIBLINEAR (Hsieh et al., 2008) can be used for training. A common strategy to train a SVM classifier with multiple settings of asymmetric costs is to reassign costs, and retrain. Our DPF method exploits the piecewise linearity property of SVM dual solution, and finds the entire solution path in one shot. This allows us to quickly construct a classifier for any choice of misclassification costs in the post-training phrase.

Quantile regression as an important statistical tool (Koenker and Bassett, 1978) has recently received attention from machine learning community. Takeuchi et al. (2006) showed that a quantile regression problem can be cast as a regularized risk minimization problem:

$$\min \frac{\lambda}{2}\|\boldsymbol{w}\|^2 + \sum_{i=1}^n \max(\tau(y_i - f_i), (1-\tau)(f_i - y_i)),$$

where $\tau \in (0,1)$ and $f_i = \boldsymbol{w}^\top \boldsymbol{x}_i$. This regression problem is very reminiscent of the quantile classification problem (3) we considered in this paper. In fact, by following the same principle as discussed in Section 4 we can extend our DPF method to find quantile regres-sion solutions for all choices of $\tau$.

Table 1: Datasets and regularization constants $\lambda$.

| Dataset | Tr./Val./Te. | Dim. | Spa. | $\lambda$ |
|---------|--------------|------|------|-----------|
| Diabetes | 668/50/50 | 8 | 13.4 % | $10^{-3}$ |
| Spam | 2500/2500/2500 | 48784 | 99.7 % | $10^{-6}$ |
| Splice | $10^4/10^4/10^4$ | 804 | 75.0 % | $10^{-2}$ |

## 6   Experiments

We now evaluate the generalization performance of a quantile classifier for various values of $\tau$, and compare the time complexity of our DPF method (Algorithm 1) with a state-of-the-art linear SVM classifier: LIBLINEAR version 1.32 (Hsieh et al., 2008). We used the LBFGSB quasi-Newton method of Byrd et al. (1995) to solve the dual problem at Step 3 of the Algorithm 1; and the conjugate gradient method was applied to find $\Delta\boldsymbol{\alpha}_{\mathcal{M}}^\tau$ at Step 13. We ran DPF without caching the $Q$ matrix.

Our experiments used three datasets: the UCI diabetes dataset (Asuncion and Newman, 2007), the spam dataset for task A of the ECML/PKDD 2006 discovery challenge,[2] and $3 \times 10^4$ worm splice samples from a biological dataset provided by Sören Sonnenburg.[3] Table 1 summarizes the datasets and our parameter settings. In all experiments the regularization parameter was chosen from the set $10^{\{-6, -5, \cdots, -1\}}$, such that a SVM classifier with symmetric misclassification costs achieves the highest prediction accuracy on the validation set, while generalization performance is reported on the test set. We included a bias $b$ in the decision function: sign $f(\boldsymbol{x}) := \boldsymbol{w}^\top \boldsymbol{x} + b$ by using the following strategy: $\boldsymbol{x}_i \leftarrow [\boldsymbol{x}_i^\top, 1]^\top, \boldsymbol{w} \leftarrow [\boldsymbol{w}^\top, b]^\top$.

All experiments were carried out on a Linux machine with dual 2.4 GHz Intel Core 2 processors and 4 GB of RAM. Our Python code is available for download from http://users.rsise.anu.edu.au/~jinyu/Code/DPF.tar.gz.

Our first set of experiments shows the influence of the quantile $\tau$ on the behaviour of a classifier. As $\boldsymbol{\alpha}^\tau$ changes, the generalization performance of the quantile classifier in terms of TPR (a.k.a. sensitivity) and TNR (a.k.a. specificity) changes accordingly. Figure 2 shows that TPR decreases (but not necessarily strictly decreases) with $\tau$, while TNR has an opposite trend. This is because increasing $\tau$ corresponds to increasing the false positive cost $C_+$ (cf. Lemma 2.2; see also

Figure 1, right), which leads the classifier to recognize more and more instances as negative samples at an expense of a decreasing TPR. At $\tau = 0$ (resp. $\tau = 1$), the classifier simply resorts to labeling all the points as + (resp. −). Therefore at these extreme points, the prediction accuracy depends on the proportion of the positive and negative samples in the dataset. For instance, on the splice dataset where 5.5% of the data is labeled as +, we obtain 5.5% accuracy at $\tau = 0$. For intermediate values of $\tau$ the prediction accuracy depends on cleanliness of the dataset measured as the total percentage of the data which lies at the margin. For instance, on the spam dataset for $\tau = 0$, around 0.28% of the training samples were at the margin. This number stabilized to about 0.32% for $\tau \in (0.0, 0.9]$, leading to very stable classification accuracy, as can be seen from Figure 2.

Clearly, finding the solution for any value of $\tau \in [0, 1]$ is more time consuming than finding the solution for a fixed $\tau$. To investigate the excess time spent in this endeavor, we compare the time complexity of our DFP algorithm with one single run of LIBLINEAR, a state of the art linear SVM training algorithm which can handle asymmetric classification costs.[4] Our second comparator is the LBFGSB algorithm, which can also be used to train a linear SVM for any fixed value of $\tau$. The core functions of LIBLINEAR and LBFGSB are implemented in C++ and Fortran, respectively (We called these functions through their Python wrappers.), while our DPF algorithm is implemented in Python, which is inherently 2 to 5 times slower. Therefore, our CPU time comparison is in favor of LIBLINEAR and LBFGSB.

Recall that our DFP algorithm invokes any linear SVM solver to find the initial solution,[5] and then finds the solution path by constructing $\mathcal{K}$. We compute the ratio of the CPU time spent on constructing $\mathcal{K}$ to the average time required by LIBLINEAR to find a solution for a given $\tau$. The averaging is done by running LIBLINEAR (resp. LBFGSB) to compute the solution for $\tau = 0.1, 0.2, \ldots, 0.9$. As shown in Table 2 on the diabetes dataset DPF finds about $2 \times 10^3$ kinks, spending $2.3 \times 10^3$ (resp. 28) times of the average LIBLINEAR (resp. LBFGSB) running time. The running time of DPF increases to $3.6 \times 10^3$ (resp. 69) times of that required by a typical run of LIBLINEAR (resp. LBFGSB) on the splice dataset where it finds over $2 \times 10^4$ kinks. We found empirically that the number of kinks, $|\mathcal{K}|$, increases with the size of training set, $n$, but is bounded by $n \log(n)$.

---

[2] The original evaluation set (http://www.ecmlpkdd2006.org/challenge.html) was equally divided into training, validation, and test set.

[3] http://www.fml.tuebingen.mpg.de/raetsch/projects/lsmkl

[4] We called LIBLINEAR with input arguments '-s 3 -B 1 -e 1e-3 -w1 (2-2$\tau$) -w-1 (2$\tau$) -c 1/($n\lambda$)'.

[5] Although in theory this is true, in practice we find that in the extreme case of $\tau = 0$, LIBLINEAR's performance degrades dramatically. Therefore, we exclusively use LBFGSB as an initial solver

Table 2: Average CPU seconds for recovering a solution from kink information (Recover) contrasted with the average time of running `LIBLINEAR` and `LBFGSB` to find $\boldsymbol{\alpha}^\tau$ for $\tau \in \{0.1, 0.2, \cdots, 0.9\}$. We also show the time required by `DPF` for path-finding. The final column $|\mathcal{K}|$ lists the number of kinks.

| Dataset | CPU Seconds | | | | $|\mathcal{K}|$ |
|---|---|---|---|---|---|
| | Recover | LIBLINEAR | LBFGSB | DPF | |
| `Diabetes` | 0.003 | 0.004 | 0.401 | 11.104 | 1886 |
| `Spam` | 0.012 | 0.467 | 11.102 | 1106.413 | 6660 |
| `Splice` | 0.103 | 0.363 | 19.155 | 1315.878 | 22611 |

It is not surprising that `DFP` is computationally more expensive than a single run of `LIBLINEAR` and `LBFGSB`. But as can be seen in Table 2, after one run of `DFP`, we can recover the solution for **any** $\tau$ efficiently. For instance, on the `spam` dataset, this only requires 0.012 seconds, compared to 0.467 seconds (resp. 11.102 seconds) for a single run of `LIBLINEAR` (resp. `LBFGSB`).

# 7    Conclusions and Outlook

In this paper we first show that minimizing the asymmetric hinge loss will lead to a quantile classifier which is risk optimal for asymmetric misclassification costs. We then present an algorithm which finds the entire solution path of a quantile classifier in a principled way. Given the entire solution path, we can construct a classifier for any given asymmetric classification cost very efficiently. Admittedly, our numerical experiments are preliminary. Running conjugate gradient repeatedly to find $\boldsymbol{\alpha}^\tau_{\mathcal{M}}$ is the main bottleneck in our `DFP` algorithm. We are exploring decomposition methods, which can take advantage of warm starts to reduce the computational burden. Future work includes extension of our algorithm to quantile regression and to multi-class classification problems.

### Acknowledgements

NICTA is funded by the Australian Government's Backing Australia's Ability and the Centre of Excellence programs. This work is also supported by the IST Program of the European Community, under the FP7 Network of Excellence, ICT-216886-NOE.

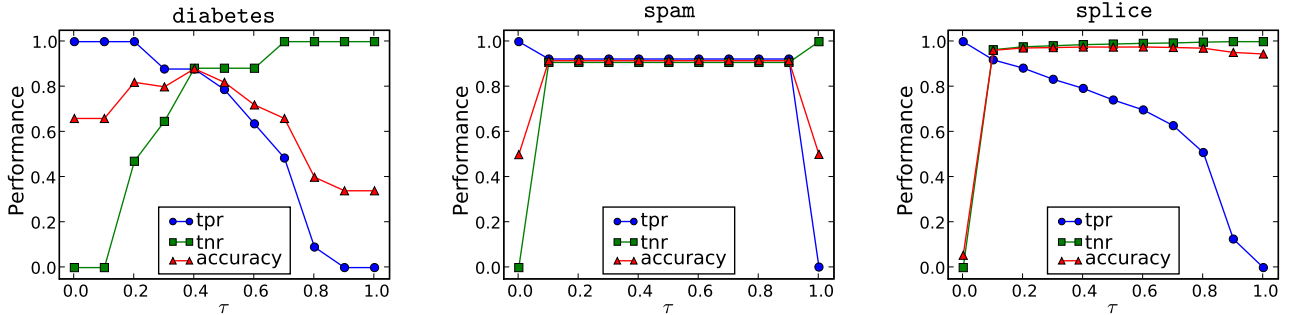

Figure 2: Our `DFP` algorithm is able to recover the solution for all values of $\tau$, which allows us to plot the true positive rate, true negative rate, and accuracy on the test dataset.

## A    Proof of Lemma 2.2

**Proof** Let $L_{\boldsymbol{x}}(f)$ be the risk conditioned on $\boldsymbol{X} = \boldsymbol{x}$:

$$L_{\boldsymbol{x}}(f) = \mathbb{E}\left[((1+Y)/2 - \tau Y)(1 - Y f(\boldsymbol{X}))_+ \mid \boldsymbol{X} = \boldsymbol{x}\right] = \tau(1 - \eta(\boldsymbol{x}))(1 + f(\boldsymbol{x}))_+ + (1 - \tau)\eta(\boldsymbol{x})(1 - f(\boldsymbol{x}))_+,$$

then we only need to show that $f^*(\boldsymbol{x})$ minimizes $L_{\boldsymbol{x}}(f)$ for any fixed $\boldsymbol{x}$.

We first show that if $\eta(\boldsymbol{x}) < \tau$ then the minimizer $f^*$ satisfies $f^*(\boldsymbol{x}) = -1$. Suppose not, that is, there exists $\boldsymbol{x}_0$ such that $\eta(\boldsymbol{x}_0) < \tau$ but $f^*(\boldsymbol{x}_0) \neq -1$. Let $\tilde{f}(\boldsymbol{x})$ be the same as $f^*(\boldsymbol{x})$ except that $\tilde{f}(\boldsymbol{x}_0) = -1$. Using the shorthand $f^*(\boldsymbol{x}_0) = f^*$, $\tilde{f}(\boldsymbol{x}_0) = \tilde{f}$ and $\eta(\boldsymbol{x}_0) = \eta$, we obtain

$$L_{\boldsymbol{x}_0}(f^*) = \tau(1 - \eta)(1 + f^*)_+ + (1 - \tau)\eta(1 - f^*)_+$$
$$\geq (1 - \tau)\eta[(1 - f^*)_+ + (1 + f^*)_+]$$
$$\geq 2(1 - \tau)\eta = L_{\boldsymbol{x}_0}(\tilde{f}),$$

where the last inequality comes from Jensen's inequality since $(.)_+$ is a convex function. For the first inequality the bound is achieved only if $f^* \leq -1$; and for the second inequality the bound is achieved only if $f^* \in [-1, 1]$. Thus when $f^* \neq -1$ it leads to a contradiction. A symmetric argument can be used to show that if $\eta(\boldsymbol{x}) > \tau$ then $f^*(\boldsymbol{x}) = 1$.    ∎

## B    KKT Optimality Conditions

The Lagrangian of the constrained optimization problem (8) takes the form of

$$L(\boldsymbol{\alpha}, \boldsymbol{\beta}, \boldsymbol{\gamma}) := D(\boldsymbol{\alpha}) - \sum_{i=1}^{n} \beta_i \alpha_i + \sum_{i=1}^{n} \gamma_i(\alpha_i - c_{y_i}^{\tau}),$$

where $\beta_i$ and $\gamma_i$ are non-negative Lagrange multipliers. The KKT conditions (Nocedal and Wright, 1999) suggest that at optimum $(\boldsymbol{\alpha}^{\tau}, \boldsymbol{\beta}^*, \boldsymbol{\gamma}^*)$ we have

$$\nabla_i L(\boldsymbol{\alpha}^{\tau}, \boldsymbol{\beta}^*, \boldsymbol{\gamma}^*) = \nabla_i D(\boldsymbol{\alpha}^{\tau}) - \beta_i^* + \gamma_i^* = 0,$$
$$\beta_i^* \alpha_i^{\tau} = 0,$$
$$\gamma_i^*(\alpha_i^{\tau} - c_{y_i}^{\tau}) = 0,$$
$$0 \leq \alpha_i^{\tau} \leq c_{y_i}^{\tau}, \quad \beta_i^* \geq 0, \quad \gamma_i^* \geq 0, \quad \forall i.$$

Simple analysis reveals that the above KKT optimality conditions constrain $\boldsymbol{\alpha}^{\tau}$ to take the form given in (12).

## C    Proof of Proposition 3.1

**Proof** Suppose the index sets (11) of $\boldsymbol{\alpha}^{\tau}$ remain unchanged for all $\tau \in (\tau_k, \tau_{k+1})$. The linearity of $\boldsymbol{\alpha}^{\tau}$ in $(\tau_k, \tau_{k+1})$ follows directly from (17). Let $\epsilon = \tau_{k+1} - \tau$, compute $\boldsymbol{\alpha}^{\tau+\epsilon}$ from $\boldsymbol{\alpha}^{\tau}$ via (17), and let $\tau_{k+1}$ be chosen in such a way that the membership of an index $i$ changes at $\boldsymbol{\alpha}^{\tau+\epsilon}$. This can only happen when $\alpha_i^{\tau+\epsilon}$ takes its boundary values: 0 or $c_{y_i}^{\tau+\epsilon}$ with $\nabla_i D(\boldsymbol{\alpha}^{\tau+\epsilon}) = 0$, which means either an $i \in \mathcal{M}$ is about to leave $\mathcal{M}$, or an $i \notin \mathcal{M}$ just moves into $\mathcal{M}$, where $\mathcal{M}$ is the margin index set of $\boldsymbol{\alpha}^{\tau}$. We now show that $\boldsymbol{\alpha}^{\tau+\epsilon}$ is optimal. To show this, we only need to show $\alpha_i^{\tau+\epsilon}$ is optimal. By construction $\nabla_i D(\boldsymbol{\alpha}^{\tau+\epsilon}) = 0$, and since $\alpha_i^{\tau+\epsilon}$ only takes 0 or $c_{y_i}^{\tau+\epsilon}$, the KKT optimality conditions (Appendix B) can be easily satisfied with appropriate choices of $\beta_i^*$ and $\gamma_i^*$, implying that $\alpha_i^{\tau+\epsilon}$ is optimal. Hence, $\boldsymbol{\alpha}^{\tau+\epsilon}$ is optimal. Therefore, we can set $\boldsymbol{\alpha}^{\tau_{k+1}} = \boldsymbol{\alpha}^{\tau+\epsilon}$, and use it as a starting point to construct subsequent dual solution path via (17). The dual solution path explored in this way is clearly continuous in $\tau$.    ∎